# Model Selection for Gaussian Process Regression by Approximation Set Coding

**Benjamin Fischer*, Nico Gorbach*, Stefan Bauer, Yatao Bian, Joachim M. Buhmann**
{FISCHEBE, NGORBACH, BAUERS, YBIAN, JBUHMANN}@INF.ETHZ.CH
*Department of Computer Science, ETH Zurich*
*8092 Zurich, Switzerland*

## Abstract

Gaussian processes are powerful, yet analytically tractable models for supervised learning. A Gaussian process is characterized by a mean function and a covariance function (kernel), which are determined by a model selection criterion. The functions to be compared do not just differ in their parametrization but in their fundamental structure. It is often not clear which function structure to choose, for instance to decide between a squared exponential and a rational quadratic kernel. Based on the principle of approximation set coding, we develop a framework for model selection to rank kernels for Gaussian process regression. In our experiments approximation set coding shows promise to become a model selection criterion competitive with maximum evidence (also called marginal likelihood) and leave-one-out cross-validation.

## 1. Introduction

A Gaussian process generalizes the multivariate Gaussian distribution to a distribution over functions. Model selection aims to adapt this distribution to a given set of data points, finding a trade-off between underfitting and overfitting. For Gaussian processes, we wish to select a mean function and a covariance function (also known as a kernel). Selecting a function is a hard problem because the possibilities are virtually unlimited. Typically, one considers a handful of function structures parametrized by hyperparameters, which are determined during model selection. Table 1 gives examples of kernels. In domains such as systems biology (Zhu et al., 2012), there is often no prior knowledge for selecting a certain function structure. A model selection criterion that is good at both hyperparameter optimization and function structure selection is thus extremely desirable.

| Name | $k(\boldsymbol{x}, \boldsymbol{x}')$ | Hyperparameters |
|---|---|---|
| Squared exponential | $\sigma_f^2 \exp\left(-\frac{1}{2\ell^2}\left\|\boldsymbol{x}-\boldsymbol{x}'\right\|_2^2\right)$ | $\ell, \sigma_f$ |
| Rational quadratic | $\sigma_f^2 \left(1+\frac{1}{2\alpha\ell^2}\left\|\boldsymbol{x}-\boldsymbol{x}'\right\|_2^2\right)^{-\alpha}$ | $\ell, \sigma_f, \alpha$ |
| Exponential | $\sigma_f^2 \exp\left(-\frac{1}{\ell}\left\|\boldsymbol{x}-\boldsymbol{x}'\right\|_2\right)$ | $\ell, \sigma_f$ |
| Periodic | $\sigma_f^2 \exp\left(-\frac{2}{\ell^2}\sin^2\left(\frac{\pi}{T}\left\|\boldsymbol{x}-\boldsymbol{x}'\right\|_2\right)\right)$ | $\ell, T, \sigma_f$ |

Table 1: Examples of kernel structures with their hyperparameters (Rasmussen and Williams, 2006).

$*$. The first two authors contributed equally to this work

### 1.1 Existing model selection criteria for Gaussian processes

Two well-known model selection criteria for Gaussian processes exist, namely maximum evidence and cross validation. Maximum evidence (also called marginal likelihood) maximizes the probability of the data under the model assumptions. Given a regression data set of inputs $\boldsymbol{X}$ with corresponding outputs $\boldsymbol{y}$, the objective of maximum evidence is to maximize the evidence $p_{\boldsymbol{\theta}}\left(\boldsymbol{y} \mid \boldsymbol{X}\right)$ with respect to the hyperparameter vector $\boldsymbol{\theta}$. We write $p_{\boldsymbol{\theta}}\left(\cdot\right)$ to make clear that this probability density function depends on $\boldsymbol{\theta}$. Cross-validation, on the other hand, minimizes an estimated generalization error of the model. In case of $K$-fold cross-validation, the objective is $-\frac{1}{K}\sum_{k=1}^{K}\log p_{\boldsymbol{\theta}}\left(\boldsymbol{y}_k \mid \boldsymbol{X}, \boldsymbol{y}_{-k}\right)$, where the outputs $\boldsymbol{y}_k$ of the $k^{\text{th}}$ fold are used to validate the model trained on the remaining outputs $\boldsymbol{y}_{-k}$. A lesser-known criterion is to minimize a bound on the generalization error from the framework of probably approximately correct (PAC) learning. However, while the concerned PAC-Bayesian theorem holds for Gaussian process classification, it seems unclear whether it can be applied to Gaussian process regression (Seeger, 2002).

Maximum evidence is generally preferred "if you really trust your prior," (Chapelle, 2005, p. 19) for instance, if one is sure about the choice of the kernel structure, so that only its hyperparameters need to be optimized. Under certain circumstances, cross-validation is more resistant to model misspecification. However, it may suffer from a higher variance (Bachoc, 2013) which is a potential risk for model selection (Cawley and Talbot, 2010). These criteria (including ours) can be used for model evaluation in automatic model construction (Lloyd et al., 2014).

### 1.2 Our contribution

Approximation set coding (ASC) determines an optimal trade-off between the expressiveness of a model and the reproducibility of its inference (Buhmann, 2010, 2013). A first contribution of our work is to extend approximation set coding to any models that define a parameter prior and a likelihood, as is the case for Bayesian linear regression. This gives birth to a family of model selection algorithms.

Second, the developed framework is applied to Gaussian process regression, which naturally comes with a prior and a likelihood. The resulting model selection criterion is then compared to the classics of maximum evidence and leave-one-out (LOO) cross-validation on the two distinct sub-problems of hyperparameter optimization and function structure selection. Although approximation set coding does not clearly perform best in our experiments, there are cases where it competes with the classics. In an experiment for kernel structure selection based on real-world data, it is interesting to see how maximum evidence and leave-one-out cross-validation disagree on which kernel structure should model the data best. This demonstrates the difficulty of model selection, but also raises the question about how to combine multiple criteria into a unifying one.

## 2. Approximation set coding for Gaussian process regression

In this section we first introduce the general model selection framework, then explain how to apply it to model selection for Gaussian process regression.

### 2.1 General model selection framework using approximation set coding

At the core of approximation set coding (Buhmann, 2010) is the idea to map the problem of model selection to an imaginary communication scenario. The principle is then to minimize an upper bound on the communication error with respect to the hyperparameters $\boldsymbol{\theta}$. Approximation set coding has been used for a variety of applications, for example, selecting the number of clusters in $K$-means clustering (Chehreghani et al., 2012), selecting the rank for a truncated singular value decomposition (Frank and Buhmann, 2011), and determining the optimal early stopping time (Gronskiy and Buhmann, 2014; Bian et al., 2015) in the algorithmic regularization framework.

More specifically, the algorithm for model selection randomly partitions a given data set $\mathcal{D}$ into two subsets $\mathcal{D}_1$ and $\mathcal{D}_2$. Let $\boldsymbol{\alpha} \in \mathbb{R}^M$ denote a parameter vector in our model – in linear regression, these would be the coefficients. Assuming that we have a likelihood $p_{\boldsymbol{\theta}}\left(\mathcal{D} \mid \boldsymbol{\alpha}\right)$ and a prior $p_{\boldsymbol{\theta}}\left(\boldsymbol{\alpha}\right)$, approximation set coding optimizes the hyperparameters by

$$\boldsymbol{\theta}^\star \in \arg\max_{\boldsymbol{\theta}} \eta_{\boldsymbol{\theta}} \qquad \text{where} \tag{1}$$
$$\eta_{\boldsymbol{\theta}} = \int_{\mathbb{R}^M} p_{\boldsymbol{\theta}}\left(\boldsymbol{\alpha} \mid \mathcal{D}_1\right) p_{\boldsymbol{\theta}}\left(\boldsymbol{\alpha} \mid \mathcal{D}_2\right) p_{\boldsymbol{\theta}}\left(\boldsymbol{\alpha}\right) \, \mathrm{d}^M \boldsymbol{\alpha}.$$

The expression $\eta_{\boldsymbol{\theta}}$ is called posterior agreement since it measures how much the posteriors $p_{\boldsymbol{\theta}}\left(\boldsymbol{\alpha} \mid \mathcal{D}_i\right)$ overlap. The prior $p_{\boldsymbol{\theta}}\left(\boldsymbol{\alpha}\right)$ gives more weight to the agreement corresponding to parameters that are a priori more plausible. An example is illustrated by Figure 1. One can consider $J$ partitions of the data into subsets $\mathcal{D}_i$, averaging their $\eta_{\boldsymbol{\theta}}$ to improve the estimate for the error bound.

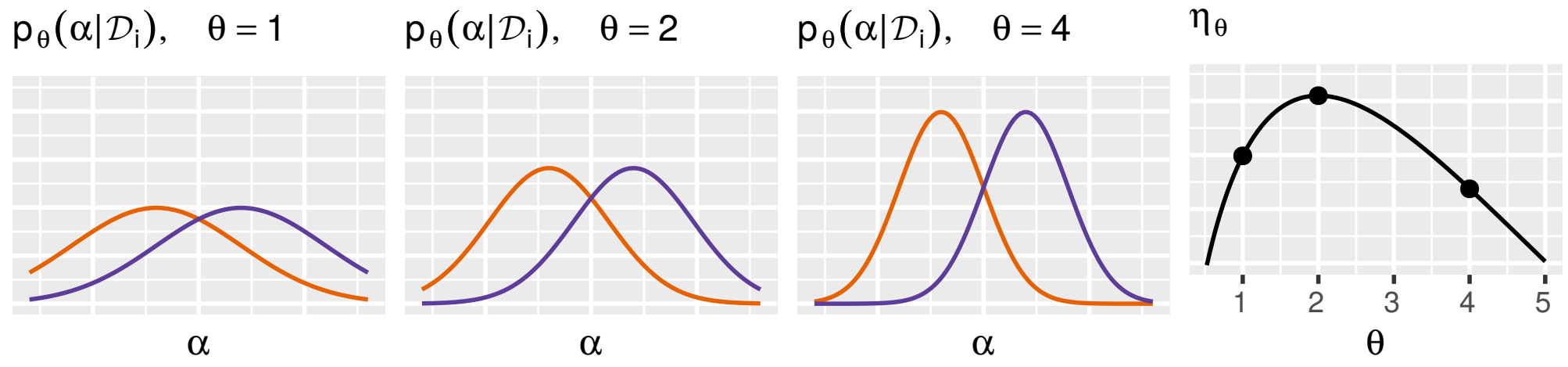

Figure 1: Posteriors $p_\theta\left(\alpha \mid \mathcal{D}_1\right)$ and $p_\theta\left(\alpha \mid \mathcal{D}_2\right)$ over the parameters $\alpha$ for various hyperparameters $\theta$. Approximation set coding basically maximizes the posterior agreement $\eta_\theta$ of Equation (1), so that in this example it would select a model with $\theta = 2$.

There are several variants to infer a posterior. The first one is via Bayes' theorem, that is, $p_{\boldsymbol{\theta}}\left(\boldsymbol{\alpha} \mid \mathcal{D}\right) \propto p_{\boldsymbol{\theta}}\left(\mathcal{D} \mid \boldsymbol{\alpha}\right) p_{\boldsymbol{\theta}}\left(\boldsymbol{\alpha}\right)$. We refer to the corresponding model selection criterion as *Bayesian approximation set coding*. The maximum entropy principle (Jaynes, 1957a,b) is a second variant rooted in the original formulation of approximation set coding (Buhmann, 2013) that determines a posterior which requires the least a priori commitment to any particular hypothesis. Using the negative log-likelihood $-\log p_{\boldsymbol{\theta}}\left(\mathcal{D} \mid \boldsymbol{\alpha}\right)$ as the cost function , the maximum entropy posterior is given by,

$$p_{\boldsymbol{\theta}}\left(\boldsymbol{\alpha} \mid \mathcal{D}\right) \propto \exp[-\beta(-\log p_{\boldsymbol{\theta}}\left(\mathcal{D} \mid \boldsymbol{\alpha}\right))] = p_{\boldsymbol{\theta}}\left(\mathcal{D} \mid \boldsymbol{\alpha}\right)^\beta$$

where $\beta$ is the inverse temperature that controls the width of the distribution. The precision $\beta$ is to be optimized alongside the hyperparameters $\boldsymbol{\theta}$. Since $\beta$ controls the noise level alongside parameters of the likelihood $\sigma_\mathrm{n}$, we fix $\beta = 1$ (called *$\beta$-noise approximation set coding*), thereby assigning $\sigma_\mathrm{n}$ the (inverse) role of $\beta$.

### 2.2 Application to Gaussian process regression

Given $N$ inputs arranged as columns of a matrix $\boldsymbol{X} \in \mathbb{R}^{D\times N}$, a Gaussian process defines a joint distribution over the corresponding $N$ function values arranged as a vector $\boldsymbol{f} \in \mathbb{R}^N$. The distribution over $\boldsymbol{f}$ is an $N$-dimensional Gaussian $\boldsymbol{f} \mid \boldsymbol{X} \sim \mathcal{N}(m(\boldsymbol{X}), k(\boldsymbol{X}, \boldsymbol{X}))$, where the mean function $m(\cdot)$ and the kernel $k(\cdot,\cdot)$ characterize the Gaussian process. In Gaussian process regression, the outputs $\boldsymbol{y} \in \mathbb{R}^N$ are modeled by the latent $\boldsymbol{f}$ affected by noise: the likelihood is $\boldsymbol{y} \mid \boldsymbol{f} \sim \mathcal{N}\left(\boldsymbol{f}, \sigma_{\mathrm{n}}^2 \boldsymbol{I}\right)$ for the noise level $\sigma_{\mathrm{n}}$.

To apply approximation set coding, the parameters $\boldsymbol{\alpha}$ are given by random subvectors of $\boldsymbol{f}$ with $M \leq N$ entries. The hyperparameters $\boldsymbol{\theta}$ are those of the mean function and the kernel as well as $\sigma_{\mathrm{n}}$. The resulting posterior agreements $\eta_{\boldsymbol{\theta}}$ for *Bayesian* and *$\beta$-noise approximation set coding* can be calculated analytically. The time complexity to all the criteria is $\Theta\left(N^3\right)$, asymptotically on a par with the objectives of maximum evidence and leave-one-out cross-validation.

For an $M$-dimensional posterior agreement, let $\widetilde{\boldsymbol{X}} \in \mathbb{R}^{D\times M}$ be made of $M$ distinct columns of $\boldsymbol{X}$, with the corresponding latent function values being $\widetilde{\boldsymbol{f}} \in \mathbb{R}^M$. The data is randomly partitioned into two subsets denoted by $(\boldsymbol{X}_1, \boldsymbol{y}_1)$ and $(\boldsymbol{X}_2, \boldsymbol{y}_2)$. For notational convenience we further define

$$
\begin{aligned}
\widetilde{\boldsymbol{m}} &= m\left(\widetilde{\boldsymbol{X}}\right), & \widetilde{\boldsymbol{K}} &= k\left(\widetilde{\boldsymbol{X}}, \widetilde{\boldsymbol{X}}\right), \\
\boldsymbol{m}_i &= m(\boldsymbol{X}_i), & \boldsymbol{K}_i &= k(\boldsymbol{X}_i, \boldsymbol{X}_i) + \sigma_{\mathrm{n}}^2 \boldsymbol{I}, \\
\widetilde{\boldsymbol{K}}_i &= k\left(\boldsymbol{X}_i, \widetilde{\boldsymbol{X}}\right), & &
\end{aligned}
$$

for $i = 1, 2$. The predictive distribution is given by

$$
\begin{bmatrix} \widetilde{\boldsymbol{f}} \\ \boldsymbol{y}_i \end{bmatrix} \sim \mathcal{N}\left( \begin{bmatrix} \widetilde{\boldsymbol{m}} \\ \boldsymbol{m}_i \end{bmatrix}, \begin{bmatrix} \widetilde{\boldsymbol{K}} & \widetilde{\boldsymbol{K}}_i^{\mathsf{T}} \\ \widetilde{\boldsymbol{K}}_i & \boldsymbol{K}_i \end{bmatrix} \right). \tag{2}
$$

For this Gaussian distribution to be non-degenerate, $\widetilde{\boldsymbol{K}}$ needs to be symmetric positive-definite. Hence, we constrain the choice of $\widetilde{\boldsymbol{X}}$ such that $\widetilde{\boldsymbol{K}}$ is invertible. The derivations for the variants of the posterior agreement $\eta_{\boldsymbol{\theta}}$ will need propositions about Gaussian distributions, which are deferred to Appendix A for ease of reading.

***Bayesian ASC.*** To derive the formula, let

$$
\begin{aligned}
\boldsymbol{s} &= \boldsymbol{V}_1^{-1}\boldsymbol{s}_1 + \boldsymbol{V}_2^{-1}\boldsymbol{s}_2 + \widetilde{\boldsymbol{K}}^{-1}\widetilde{\boldsymbol{m}}, & \boldsymbol{B}_i &= \boldsymbol{K}_i^{-1}\widetilde{\boldsymbol{K}}_i, \\
\boldsymbol{s}_i &= \widetilde{\boldsymbol{m}} + \boldsymbol{B}_i^{\mathsf{T}}(\boldsymbol{y}_i - \boldsymbol{m}_i), & \boldsymbol{V}_i &= \widetilde{\boldsymbol{K}} - \widetilde{\boldsymbol{K}}_i^{\mathsf{T}}\boldsymbol{B}_i, \\
\boldsymbol{P} &= \boldsymbol{V}_1^{-1} + \boldsymbol{V}_2^{-1} + \widetilde{\boldsymbol{K}}^{-1}. & &
\end{aligned}
$$

Applying Proposition 1 to Equation (2), the posterior evaluated at $\widetilde{\boldsymbol{X}}$ is

$$
\widetilde{\boldsymbol{f}} \,\Big|\, \boldsymbol{X}_i, \boldsymbol{y}_i, \widetilde{\boldsymbol{X}} \sim \mathcal{N}(\boldsymbol{s}_i, \boldsymbol{V}_i).
$$

Since $\widetilde{\boldsymbol{f}} \sim \mathcal{N}\left(\widetilde{\boldsymbol{m}}, \widetilde{\boldsymbol{K}}\right)$, and according to Equation (1), the posterior agreement of *Bayesian approximation set coding* is thus

$$
\eta_{\boldsymbol{\theta}}^{\text{Bayesian}} = \int_{\mathbb{R}^M} \mathcal{N}\left(\widetilde{\boldsymbol{f}} \,\middle|\, \boldsymbol{s}_1, \boldsymbol{V}_1\right) \mathcal{N}\left(\widetilde{\boldsymbol{f}} \,\middle|\, \boldsymbol{s}_2, \boldsymbol{V}_2\right) \mathcal{N}\left(\widetilde{\boldsymbol{f}} \,\middle|\, \widetilde{\boldsymbol{m}}, \widetilde{\boldsymbol{K}}\right) \mathrm{d}^M \widetilde{\boldsymbol{f}}.
$$

Using Proposition 3, it amounts to

$$\eta_{\boldsymbol{\theta}}^{\text{Bayesian}} = \frac{\mathcal{N}\left(\boldsymbol{s}_1 \mid \mathbf{0}, \boldsymbol{V}_1\right)\mathcal{N}\left(\boldsymbol{s}_2 \mid \mathbf{0}, \boldsymbol{V}_2\right)}{|\boldsymbol{P}|\mathcal{N}\left(\boldsymbol{s} \mid \mathbf{0}, \boldsymbol{P}\right)}\mathcal{N}\left(\widetilde{\boldsymbol{m}} \,\middle|\, \mathbf{0}, \widetilde{\boldsymbol{K}}\right).$$

Even though it is in closed form, it is generally non-convex.

***β-noise approximation set coding*** Let

$$\begin{aligned}
\boldsymbol{r} &= \boldsymbol{r}_1 + \boldsymbol{r}_2 + \widetilde{\boldsymbol{K}}^{-1}\widetilde{\boldsymbol{m}}, & \boldsymbol{\Lambda} &= \boldsymbol{\Lambda}_1 + \boldsymbol{\Lambda}_2 + \widetilde{\boldsymbol{K}}^{-1},\\
\boldsymbol{r}_i &= \boldsymbol{A}_i\boldsymbol{\Sigma}_i^{-1}\boldsymbol{\mu}_i, & \boldsymbol{\Lambda}_i &= \boldsymbol{A}_i\boldsymbol{\Sigma}_i^{-1}\boldsymbol{A}_i^{\intercal},\\
\boldsymbol{\mu}_i &= \boldsymbol{y}_i - \boldsymbol{m}_i + \boldsymbol{A}_i^{\intercal}\widetilde{\boldsymbol{m}}, & \boldsymbol{\Sigma}_i &= \boldsymbol{K}_i - \widetilde{\boldsymbol{K}}_i\boldsymbol{A}_i,\\
\boldsymbol{A}_i &= \widetilde{\boldsymbol{K}}^{-1}\widetilde{\boldsymbol{K}}_i^{\intercal}.
\end{aligned}$$

Again applying Proposition 1 to Equation (2),

$$\boldsymbol{y}_i \,\middle|\, \widetilde{\boldsymbol{X}}, \widetilde{\boldsymbol{f}}, \boldsymbol{X}_i \sim \mathcal{N}\left(\boldsymbol{m}_i + \boldsymbol{A}_i^{\intercal}\left(\widetilde{\boldsymbol{f}} - \widetilde{\boldsymbol{m}}\right), \boldsymbol{\Sigma}_i\right).$$

The corresponding density can be rewritten as

$$p\left(\boldsymbol{y}_i \,\middle|\, \widetilde{\boldsymbol{X}}, \widetilde{\boldsymbol{f}}, \boldsymbol{X}_i\right) = \mathcal{N}\left(\boldsymbol{A}_i^{\intercal}\widetilde{\boldsymbol{f}} \,\middle|\, \boldsymbol{\mu}_i, \boldsymbol{\Sigma}_i\right).$$

We now to derive a maximum entropy density over $\widetilde{\boldsymbol{f}}$ which yields a new posterior distribution

$$\frac{\mathcal{N}\left(\boldsymbol{A}_i^{\intercal}\widetilde{\boldsymbol{f}} \,\middle|\, \boldsymbol{\mu}_i, \boldsymbol{\Sigma}_i\right)^{\beta}}{\int_{\mathbb{R}^M}\mathcal{N}\left(\boldsymbol{A}_i^{\intercal}\widetilde{\boldsymbol{f}} \,\middle|\, \boldsymbol{\mu}_i, \boldsymbol{\Sigma}_i\right)^{\beta}\,\mathrm{d}^M\widetilde{\boldsymbol{f}}} = \mathcal{N}\left(\widetilde{\boldsymbol{f}} \,\middle|\, \boldsymbol{\Lambda}_i^{-1}\boldsymbol{r}_i, \boldsymbol{\Lambda}_i^{-1}\right),$$

where we use Proposition 6 to normalize the distribution. We further assume that $\boldsymbol{A}_i$ has full row rank and we set $\beta = 1$. Substituting the maximum entropy density into Equation (1) yields

$$\eta_{\boldsymbol{\theta}}^{\beta\text{−noise}} = \int_{\mathbb{R}^M}\mathcal{N}\left(\widetilde{\boldsymbol{f}} \,\middle|\, \boldsymbol{\Lambda}_1^{-1}\boldsymbol{r}_1, \boldsymbol{\Lambda}_1^{-1}\right)\mathcal{N}\left(\widetilde{\boldsymbol{f}} \,\middle|\, \boldsymbol{\Lambda}_2^{-1}\boldsymbol{r}_2, \boldsymbol{\Lambda}_2^{-1}\right)\mathcal{N}\left(\widetilde{\boldsymbol{f}} \,\middle|\, \widetilde{\boldsymbol{m}}, \widetilde{\boldsymbol{K}}\right)\,\mathrm{d}^M\widetilde{\boldsymbol{f}},$$

which by Proposition 3 is

$$\eta_{\boldsymbol{\theta}}^{\beta\text{−noise}} = \frac{\mathcal{N}\left(\boldsymbol{\Lambda}_1^{-1}\boldsymbol{r}_1 \mid \mathbf{0}, \boldsymbol{\Lambda}_1^{-1}\right)\mathcal{N}\left(\boldsymbol{\Lambda}_2^{-1}\boldsymbol{r}_2 \mid \mathbf{0}, \boldsymbol{\Lambda}_2^{-1}\right)}{|\boldsymbol{\Lambda}|\mathcal{N}\left(\boldsymbol{r} \mid \mathbf{0}, \boldsymbol{\Lambda}\right)}\mathcal{N}\left(\widetilde{\boldsymbol{m}} \,\middle|\, \mathbf{0}, \widetilde{\boldsymbol{K}}\right), \tag{3}$$

which is generally non-convex.

## 3. Experimental results

In our experiments, we assess *Bayesian* and *β-noise approximation set coding* for $J = 256$ data partitions with a posterior agreement of dimension $M = 2$. Any Gaussian process uses the zero mean function for simplicity. The error measure is the mean standardized log loss according to Rasmussen and Williams (2006), which considers both the predictive mean and covariance. To numerically optimize objective functions, the algorithm of limited-memory BFGS (Nocedal, 1980) is applied.

**Experiment for hyperparameter optimization on synthetic data.** First, we compare the model selection criteria on hyperparameter optimization for a fixed kernel structure. From a one-dimensional Gaussian process, we randomly draw 128 data sets, each with $N = 64$ training and 2048 test points. Every criterion is then applied to the training set to optimize the hyperparameters of a Gaussian process with the same kernel structure.

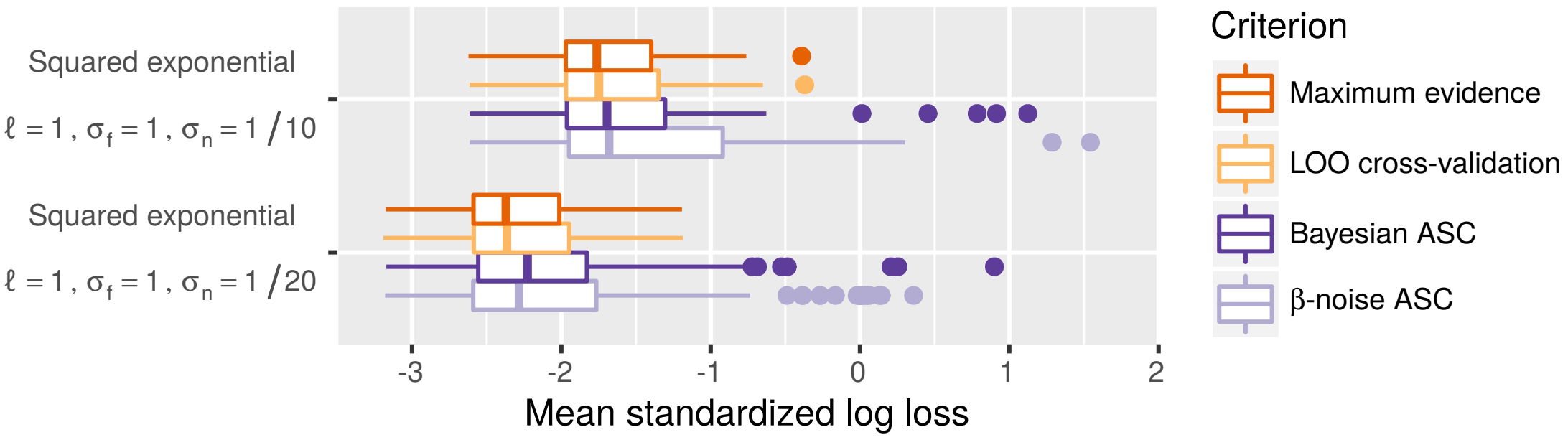

Figure 2: Test errors for hyperparameter optimization.

Figure 2 shows the test errors for the popular squared exponential kernel structure with various noise levels $\sigma_n$ (definition in Table 1). Maximum evidence generally has the best test error, which is to be expected since the kernel structure is known. It is closely followed by leave-one-out cross-validation. Both variants of approximation set coding have a worse median and a wider spread. While their outliers make them unstable, they seem to compete with the classic criteria at times.

**Experiment for kernel ranking on synthetic data.** The model selection criteria rank kernels relatively consistently for synthetic data as shown in Figure 3. Hyperparameters in the top two rows were estimated by maximum evidence whereas for the bottom two rows they were estimated by leave-one-out cross-validation. The kernel to generate the data is depicted on the right (teacher) and we fitted the squared exponential, the rational quadratic, the exponential and the periodic kernels to the data (students). In most cases the criteria selected the correct kernel that was used to generate the data. The mean rank is visualized with a 95% confidence interval. Independent of the choice for hyperparameter optimization, evidence, cross-validation and both approximation set coding based methods select the correct kernels in all four scenarios. In addition, even the confidence intervals are very similar. Overall, the periodic kernel seems to be slightly easier to learn, while the squared exponential and the rational quadratic kernel are often assigned equal ranks by all methods. Finally, all methods clearly separate the exponential kernel as a student from both the periodic and the squared exponential as teachers. Thus for kernel selection the approximation set coding based methods seem to be consistent and equally good as both evidence and cross-validation for kernel selection.

**Experiments for kernel ranking on real-world data.** Next, we compare the criteria on kernel structure selection on two public datasets for different applications.

**Berkeley Earth** As a first real-world data set, we use Earth's land temperature averaged per day of the year from 1880 until 2014[1]. It has 365 data points, which we randomly partition 256 times into $N = 64$ training and 301 test points. Given a training set, the hyperparameters are optimized by

1. http://berkeleyearth.org/data/

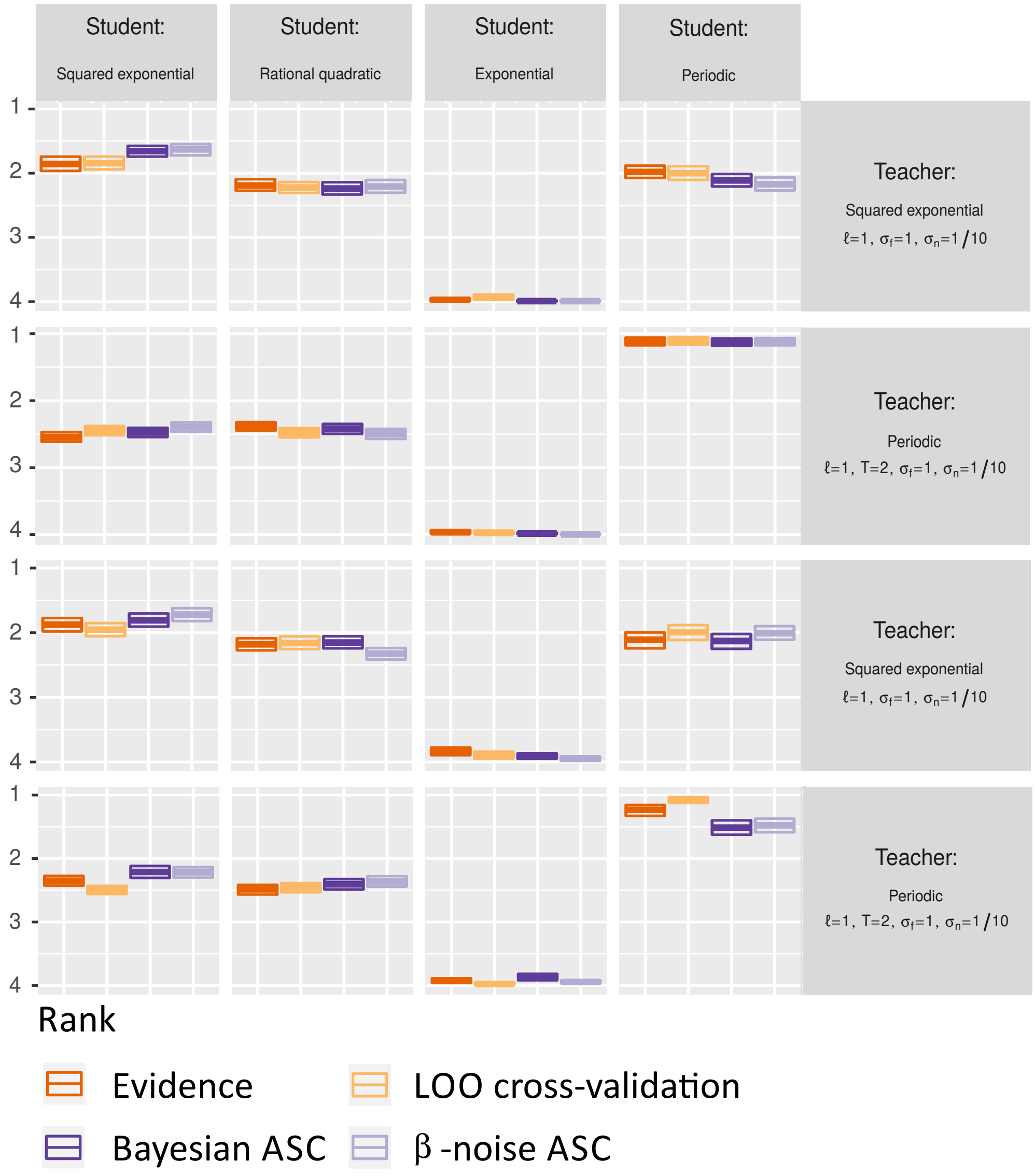

Figure 3: Ranking of kernels for synthetic data with rank 1 being the best. The top two rows estimate hyperparameters by maximum evidence and the bottom two rows by leave-one-out cross-validation. Rank 1 means the best. The mean rank is visualized with a 95% confidence interval.

leave-one-out cross-validation for each kernel structure of Table 1. The resulting Gaussian process models are then ranked by the various criteria objectives. An additional ranking according to the test error serves as a guide for the assessment.

Figure 4 shows the rankings averaged over the data partitions. According to the average test error, the exponential kernel structure seems the most suitable, followed by the rational quadratic kernel structure. Maximum evidence would select an exponential, leave-one-out cross-validation a rational quadratic, and both variants of approximation set coding a squared exponential kernel structure. It is interesting to see this clear disagreement between the criteria. An example for the learned Gaussian processes is visualized in Figure 5. The test set of this example is representative in the sense that the rankings according to the criteria and the test error are the same as on average. The exponential kernel fits a lot to the data, similar to a linear interpolation, which raises doubts about maximum evidence. Despite its unfavorable test error, the squared exponential kernel appears to be a valid choice based on this manual assessment. We conclude that approximation set coding selects a good trade-off.

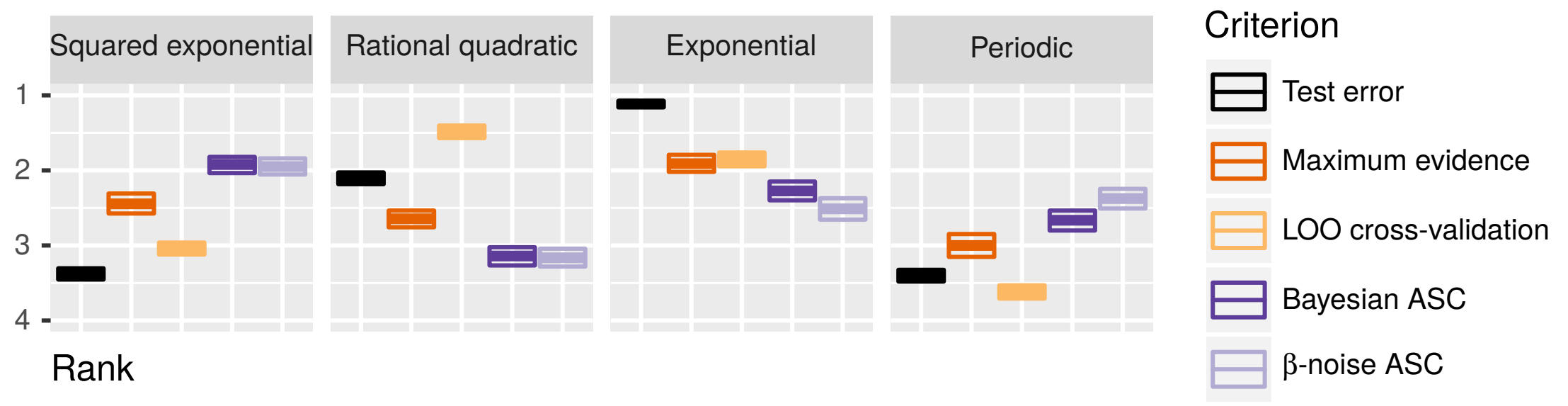

Figure 4: Kernel structure selection for Berkeley Earth's land temperature. The mean rank is visualized with a 95% confidence interval, with rank 1 being the best.

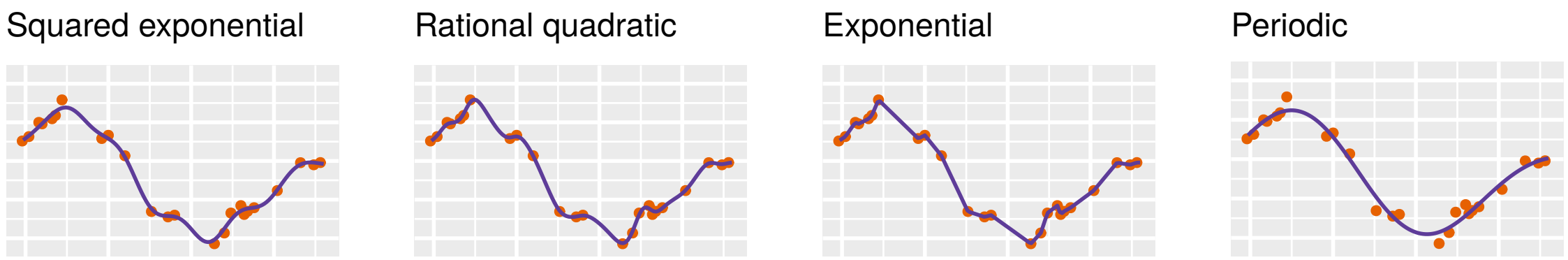

Figure 5: Predictive means (lines) for a real-world data example points from the Berkeley Earth's land temperature data.

**Combined Cycle Power Plant** The dataset contains 9568 data points collected from a Combined Cycle Power Plant over 6 years (2006–2011), when the power plant was set to work with full load[2]. Features consist of hourly average ambient variables Temperature, Ambient Pressure, Relative Humidity and Exhaust Vacuum to predict the net hourly electrical energy output of the plant. Bayesian and $\beta$-noise ASC both prefer the squared exponential kernel whereas maximum evidence prefers the periodic kernel as shown in Figure 7. The predictive means associated with the squared exponential and periodic kernels are plotted in Figure 6. For a simplified visualization, we only plotted the two most important dimensions. Given smaller variations in the other two dimensions, we conclude again that both choices, from maximum evidence or $\beta$-noise ASC seem to be valid choices.

## 4. Discussion and conclusion

In this work we developed a framework to rank kernels for Gaussian process regression and compared it to state-of-the-art methods such as maximum evidence and leave-one-out cross-validation. All

2. http://archive.ics.uci.edu/ml/datasets/Combined+Cycle+Power+Plant

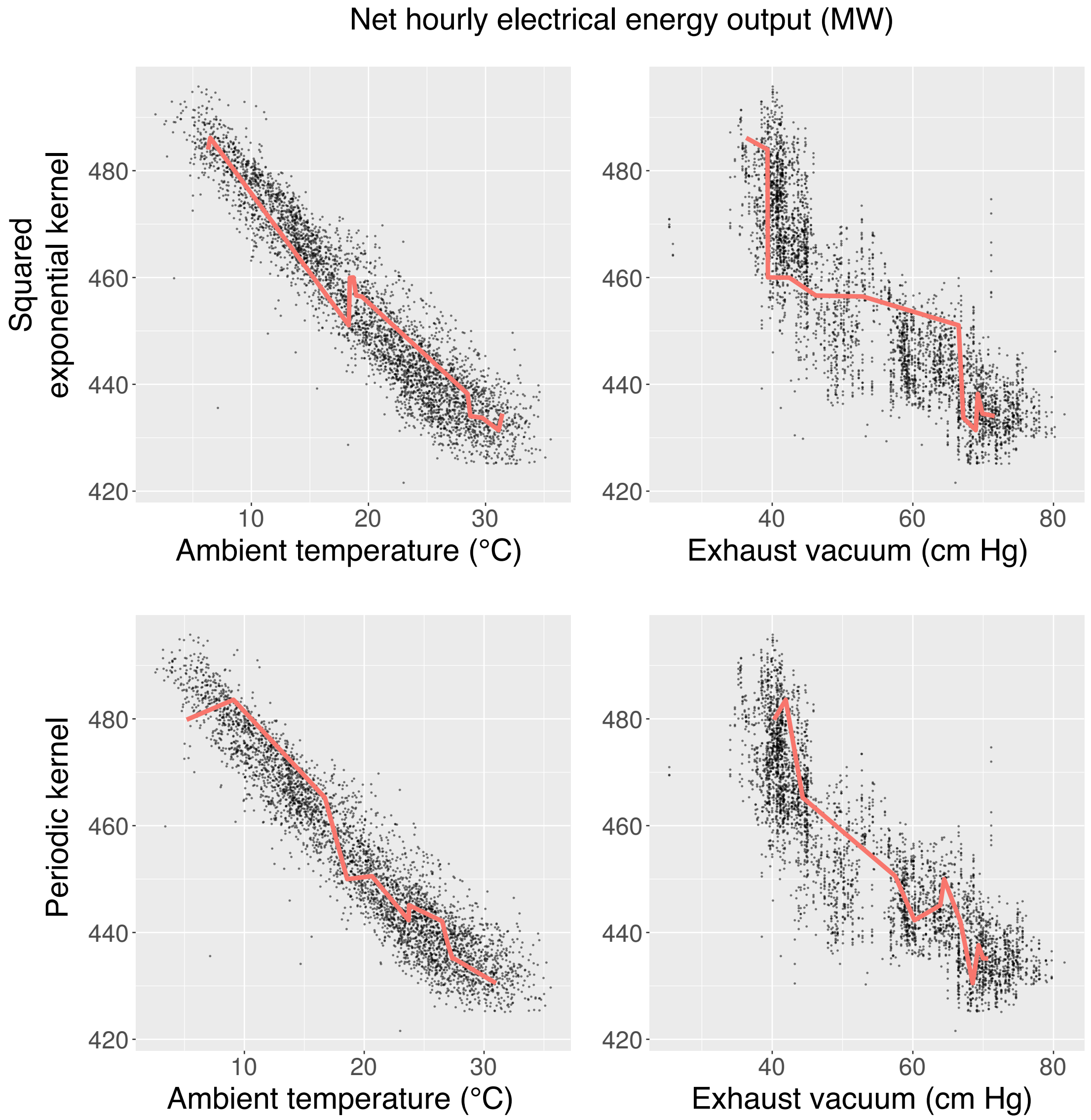

Figure 6: Test data for the net hourly electrical energy output is plotted against the ambient temperature and the exhaust vacuum. The other two dimensions, namely the ambient pressure and relative humidity are less decisive and were omitted for a simplified visualization. Bayesian and $\beta$-noise ASC both prefer the squared exponential kernel whose predictive means (red lines) is shown in the top two plots. Maximum evidence on the other hand selects the periodic kernel whose predictive means (red line) is shown in the bottom two plots.

of the objectives (Maximum evidence, Bayesian ASC and $\beta$-noise ASC) are highly non-convex functions, for which there is no efficient optimization algorithm so far to guarantee the approximation quality. This may cause some randomness in the experiments and requires future work to develop better optimization methods for these non-convex objectives. The randomness partially explains one open issue with approximation set coding. It often happens that the Cholesky decomposition of certain covariance matrices numerically fails, so that the objective cannot be evaluated. This can occur because the eigenvalues of a covariance matrix "can decay very rapidly." (Rasmussen and Williams, 2006, p. 201).

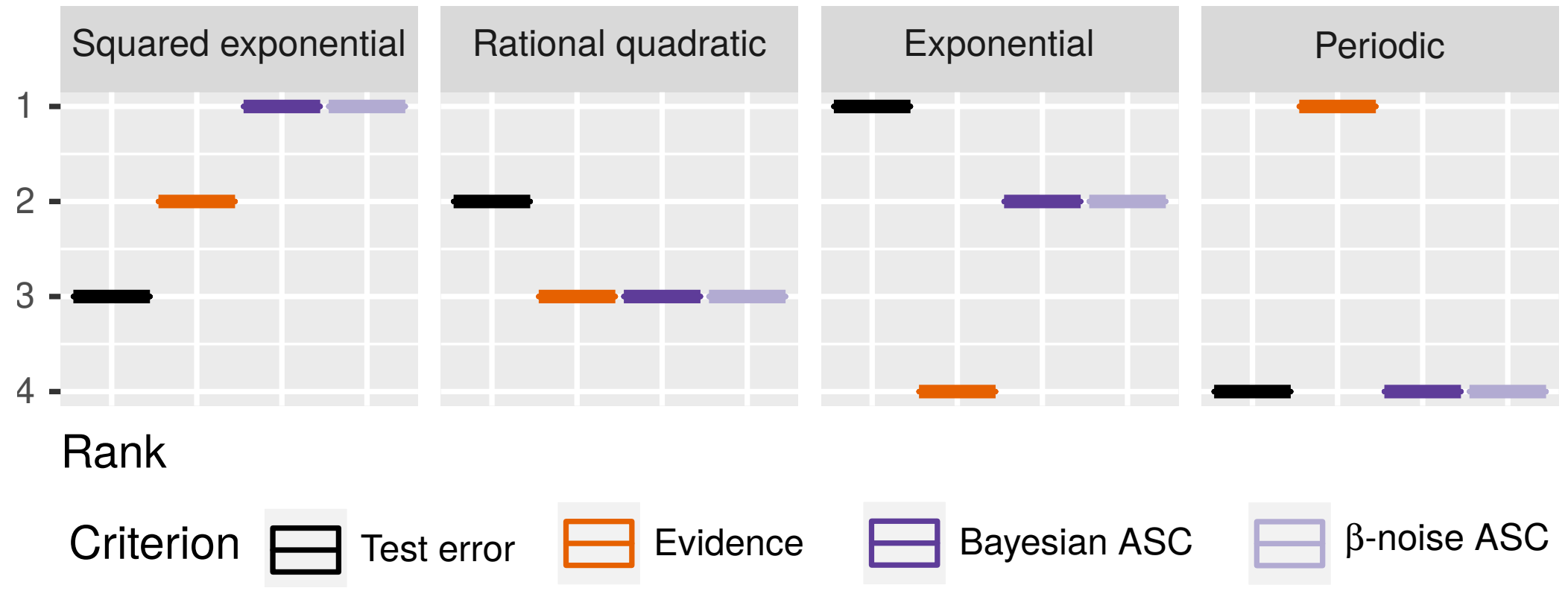

Figure 7: Ranking of kernels for the power plant data set. As before, Bayesian and $\beta$-noise consistently rank the kernels and choose the squared exponential kernel as the optimum whereas maximum evidence prefers the periodic kernel. The test error prefers the exponential kernel.

If the function structure of a Gaussian process is known, so that only its hyperparameters need to be optimized, the criterion of maximum evidence seems to perform best. However, in the usual case where the function structure is also subject to model selection, approximation set coding is a potentially better alternative according to manual examinations. We are unable to formally define what function structure should be recovered since this may possibly solve the model selection problem itself. Even though the experiments are only for Gaussian process regression, the framework is general enough to serve as a promising scheme for general model selection, e.g., GP classification and deep GP (Damianou and Lawrence, 2013).

## Acknowledgments

This research was partially supported by the Max Planck ETH Center for Learning Systems and the SystemsX.ch project SignalX.

# Appendix

## Appendix A. Propositions of Gaussian distribution

This is a collection of properties related to Gaussian distributions for the derivations in Section 2.2.

**Proposition 1** *If*

$$\begin{bmatrix} \boldsymbol{t} \\ \boldsymbol{u} \end{bmatrix} \sim \mathcal{N}\left(\begin{bmatrix} \boldsymbol{\mu} \\ \boldsymbol{r} \end{bmatrix}, \begin{bmatrix} \boldsymbol{\Sigma} & \boldsymbol{A} \\ \boldsymbol{A}^\intercal & \boldsymbol{V} \end{bmatrix}\right)$$

*then*

$$\boldsymbol{t} \mid \boldsymbol{u} \sim \mathcal{N}\left(\boldsymbol{\mu} + \boldsymbol{A}\boldsymbol{V}^{-1}\left(\boldsymbol{u} - \boldsymbol{r}\right), \boldsymbol{\Sigma} - \boldsymbol{A}\boldsymbol{V}^{-1}\boldsymbol{A}^\intercal\right)$$

*(Tong, 2012, Theorem 3.3.4).*

**Proposition 2** *If $\boldsymbol{\Lambda}$ is symmetric positive-definite, then*

$$\int_{\mathbb{R}^D} \exp\left(\boldsymbol{x}^\intercal\left(\boldsymbol{\mu} - \frac{1}{2}\boldsymbol{\Lambda}\boldsymbol{x}\right)\right) \mathrm{d}^D\boldsymbol{x} = \frac{1}{|\boldsymbol{\Lambda}|\,\mathcal{N}\left(\boldsymbol{\mu} \mid \boldsymbol{0}, \boldsymbol{\Lambda}\right)}$$

*(Zee, 2003, 14).*

**Proposition 3** *It holds that,*

$$\int_{\mathbb{R}^D} \prod_{k=1}^{K} \mathcal{N}\left(\boldsymbol{x} \mid \boldsymbol{\mu}_k, \boldsymbol{\Sigma}_k\right) \mathrm{d}^D\boldsymbol{x} = \frac{\prod_{k=1}^{K} \mathcal{N}\left(\boldsymbol{\mu}_k \mid \boldsymbol{0}, \boldsymbol{\Sigma}_k\right)}{|\boldsymbol{\Lambda}|\,\mathcal{N}\left(\boldsymbol{r} \mid \boldsymbol{0}, \boldsymbol{\Lambda}\right)},$$

*where $\boldsymbol{r} = \sum_{k=1}^{K} \boldsymbol{\Sigma}_k^{-1}\boldsymbol{\mu}_k$ and $\boldsymbol{\Lambda} = \sum_{k=1}^{K} \boldsymbol{\Sigma}_k^{-1}$.*

**Proof** We shorten $\gamma = \prod_{k=1}^{K} \mathcal{N}\left(\boldsymbol{\mu}_k \mid \boldsymbol{0}, \boldsymbol{\Sigma}_k\right)$ to move this factor $\gamma$ independent of $\boldsymbol{x}$ out of the integral as in

$$\begin{aligned}
&\int_{\mathbb{R}^D} \prod_{k=1}^{K} \mathcal{N}\left(\boldsymbol{x} \mid \boldsymbol{\mu}_k, \boldsymbol{\Sigma}_k\right) \mathrm{d}^D\boldsymbol{x} \\
&= \gamma \int_{\mathbb{R}^D} \prod_{k=1}^{K} \exp\left(\boldsymbol{x}^\intercal\boldsymbol{\Sigma}_k^{-1}\left(\boldsymbol{\mu}_k - \frac{1}{2}\boldsymbol{x}\right)\right) \mathrm{d}^D\boldsymbol{x} \\
&= \gamma \int_{\mathbb{R}^D} \exp\left(\sum_{k=1}^{K} \boldsymbol{x}^\intercal\boldsymbol{\Sigma}_k^{-1}\left(\boldsymbol{\mu}_k - \frac{1}{2}\boldsymbol{x}\right)\right) \mathrm{d}^D\boldsymbol{x} \\
&= \gamma \int_{\mathbb{R}^D} \exp\left(\boldsymbol{x}^\intercal\left(\boldsymbol{r} - \frac{1}{2}\boldsymbol{\Lambda}\boldsymbol{x}\right)\right) \mathrm{d}^D\boldsymbol{x}.
\end{aligned}$$

The remaining integral can be calculated by Proposition 2. ■

**Proposition 4** *If $\boldsymbol{\Sigma}$ is symmetric positive-definite, then $\boldsymbol{\Sigma}$ is invertible and $\boldsymbol{\Sigma}^{-1}$ is symmetric positive-definite (Horn and Johnson, 2012, 430).*

**Proposition 5** *If $\boldsymbol{\Sigma}$ is symmetric positive-definite and $\boldsymbol{A}$ has full row rank, then $\boldsymbol{A}\boldsymbol{\Sigma}\boldsymbol{A}^\intercal$ is symmetric positive-definite (Horn and Johnson, 2012, Observation 7.1.8.(b)).*

**Proposition 6** *For $\boldsymbol{A} \in \mathbb{R}^{D\times N}$ of full row rank, the density*

$$p(\boldsymbol{x}) = \frac{\mathcal{N}(\boldsymbol{A}^\intercal \boldsymbol{x} \mid \boldsymbol{\mu}, \boldsymbol{\Sigma})}{\int_{\mathbb{R}^D} \mathcal{N}(\boldsymbol{A}^\intercal \boldsymbol{x} \mid \boldsymbol{\mu}, \boldsymbol{\Sigma})\, \mathrm{d}^D \boldsymbol{x}}$$

*has the equivalent form as $p(\boldsymbol{x}) = \mathcal{N}\left(\boldsymbol{x} \mid \boldsymbol{\Lambda}^{-1}\boldsymbol{r}, \boldsymbol{\Lambda}^{-1}\right)$, where $\boldsymbol{r} = \boldsymbol{A}\boldsymbol{\Sigma}^{-1}\boldsymbol{\mu}$ and $\boldsymbol{\Lambda} = \boldsymbol{A}\boldsymbol{\Sigma}^{-1}\boldsymbol{A}^\intercal$.*

**Proof** First, we separate a factor independent of $\boldsymbol{x}$ in

$$\mathcal{N}(\boldsymbol{A}^\intercal \boldsymbol{x} \mid \boldsymbol{\mu}, \boldsymbol{\Sigma}) = \mathcal{N}(\boldsymbol{\mu} \mid \boldsymbol{0}, \boldsymbol{\Sigma}) \exp\left(\boldsymbol{x}^\intercal\left(\boldsymbol{r} - \frac{1}{2}\boldsymbol{\Lambda}\boldsymbol{x}\right)\right).$$

Therefore,

$$p(\boldsymbol{x}) = \frac{\exp\left(\boldsymbol{x}^\intercal\left(\boldsymbol{r} - \frac{1}{2}\boldsymbol{\Lambda}\boldsymbol{x}\right)\right)}{\int_{\mathbb{R}^D} \exp\left(\boldsymbol{x}^\intercal\left(\boldsymbol{r} - \frac{1}{2}\boldsymbol{\Lambda}\boldsymbol{x}\right)\right) \mathrm{d}^D \boldsymbol{x}}.$$

We now calculate the integral. From Proposition 4 and Proposition 5, one can see that $\boldsymbol{\Lambda}$ is symmetric positive-definite, so that Proposition 2 can be applied to find

$$\int_{\mathbb{R}^D} \exp\left(\boldsymbol{x}^\intercal\left(\boldsymbol{r} - \frac{1}{2}\boldsymbol{\Lambda}\boldsymbol{x}\right)\right) \mathrm{d}^D \boldsymbol{x} = \frac{1}{|\boldsymbol{\Lambda}|\,\mathcal{N}(\boldsymbol{r} \mid \boldsymbol{0}, \boldsymbol{\Lambda})}.$$

Finally, one gets

$$\begin{aligned} p(\boldsymbol{x}) &= |\boldsymbol{\Lambda}|\,\mathcal{N}(\boldsymbol{r} \mid \boldsymbol{0}, \boldsymbol{\Lambda}) \exp\left(\boldsymbol{x}^\intercal\left(\boldsymbol{r} - \frac{1}{2}\boldsymbol{\Lambda}\boldsymbol{x}\right)\right) \\ &= \mathcal{N}\left(\boldsymbol{x} \mid \boldsymbol{\Lambda}^{-1}\boldsymbol{r}, \boldsymbol{\Lambda}^{-1}\right). \end{aligned}$$

■

## References

François Bachoc. Cross validation and maximum likelihood estimations of hyper-parameters of gaussian processes with model misspecification. *Computational Statistics & Data Analysis*, 66: 55–69, 2013. doi: 10.1016/j.csda.2013.03.016. URL http://dx.doi.org/10.1016/j.csda.2013.03.016.

Yatao Bian, Alexey Gronskiy, and Joachim M. Buhmann. Greedy maxcut algorithms and their information content. In *IEEE Information Theory Workshop (ITW)*, pages 1–5, 2015.

Joachim M. Buhmann. Information theoretic model validation for clustering. In *IEEE International Symposium on Information Theory, ISIT 2010, June 13-18, 2010, Austin, Texas, USA, Proceedings*, pages 1398–1402, 2010. doi: 10.1109/ISIT.2010.5513616. URL http://dx.doi.org/10.1109/ISIT.2010.5513616.

Joachim M. Buhmann. Simbad: Emergence of pattern similarity. In Marcello Pelillo, editor, *Similarity-Based Pattern Analysis and Recognition*, Advances in Computer Vision and Pattern Recognition, page 45–64. Springer London, 2013. ISBN 978-1-4471-5627-7. doi: 10.1007/978-1-4471-5628-4_3.

Gavin C. Cawley and Nicola L. C. Talbot. On over-fitting in model selection and subsequent selection bias in performance evaluation. *Journal of Machine Learning Research*, 11:2079–2107, 2010. URL http://portal.acm.org/citation.cfm?id=1859921.

Olivier Chapelle. Some thoughts about gaussian processes. http://is.tuebingen.mpg.de/fileadmin/user_upload/files/publications/gp_[0].pdf, 2005.

Morteza Haghir Chehreghani, Alberto Giovanni Busetto, and Joachim M. Buhmann. Information theoretic model validation for spectral clustering. In *International Conference on Artificial Intelligence and Statistics (AISTATS)*, volume 22, page 495–503. Journal of Machine Learning Research, 2012. URL http://jmlr.org/proceedings/papers/v22/haghir12.html.

Andreas C. Damianou and Neil D. Lawrence. Deep gaussian processes. In *Proceedings of the Sixteenth International Conference on Artificial Intelligence and Statistics, AISTATS 2013, Scottsdale, AZ, USA, April 29 - May 1, 2013*, pages 207–215, 2013. URL http://jmlr.org/proceedings/papers/v31/damianou13a.html.

Mario Frank and Joachim M. Buhmann. Selecting the rank of truncated svd by maximum approximation capacity. In *IEEE International Symposium on Information Theory (ISIT)*, page 1036–1040. IEEE, 2011.

A. Gronskiy and J.M. Buhmann. How informative are minimum spanning tree algorithms? In *IEEE International Symposium on Information Theory (ISIT)*, pages 2277–2281, June 2014. doi: 10.1109/ISIT.2014.6875239.

Roger A. Horn and Charles R. Johnson. *Matrix Analysis*. Cambridge University Press, 2 edition, 2012. ISBN 978-0-521-83940-2.

Edwin T Jaynes. Information theory and statistical mechanics. *Physical review*, 106(4):620, 1957a.

Edwin T Jaynes. Information theory and statistical mechanics. ii. *Physical review*, 108(2):171, 1957b.

James Robert Lloyd, David Duvenaud, Roger Grosse, Joshua B. Tenenbaum, and Zoubin Ghahramani. Automatic construction and natural-language description of nonparametric regression models. In *AAAI Conference on Artificial Intelligence*, volume 28, page 1242–1250, 2014. URL http://www.aaai.org/ocs/index.php/AAAI/AAAI14/paper/view/8240.

Jorge Nocedal. Updating quasi-newton matrices with limited storage. *Mathematics of computation*, 35(151):773–782, 1980.

Carl Edward Rasmussen and Christopher K. I. Williams. *Gaussian Processes for Machine Learning*. Adaptive Computation and Machine Learning. The MIT Press, 2006. ISBN 978-0-262-18253-9.

Matthias W. Seeger. Pac-bayesian generalisation error bounds for gaussian process classification. *Journal of Machine Learning Research*, 3:233–269, 2002. URL http://www.jmlr.org/papers/v3/seeger02a.html.

Yung Liang Tong. *The multivariate normal distribution*. Springer Science & Business Media, 2012.

Anthony Zee. *Quantum Field Theory in a Nutshell*. Princeton University Press, 2003. ISBN 978-0-691-01019-9.

Xiangxin Zhu, Max Welling, Fang Jin, and John S. Lowengrub. Predicting simulation parameters of biological systems using a gaussian process model. *Statistical Analysis and Data Mining*, 5(6):509–522, 2012. doi: 10.1002/sam.11163. URL http://dx.doi.org/10.1002/sam.11163.